\documentclass{article}
\usepackage{ijcai20}

\usepackage{times}
\usepackage{soul}
\usepackage{url}
\usepackage[hidelinks]{hyperref}
\usepackage[utf8]{inputenc}
\usepackage[small]{caption}
\usepackage{graphicx}
\usepackage{amsfonts}
\usepackage{amsmath}
\usepackage{amsthm}
\usepackage{booktabs}
\usepackage{algorithm}
\usepackage{multirow}
\usepackage{algorithmic}
\title{Data Augmentation for Copy-Mechanism in Dialogue State Tracking}

\author{
Xiaohui Song$^1$\and
Liangjun Zang$^1$\and
Yipeng Su$^1$\and
Xing Wu$^3$\and
Jizhong Han$^1$ \And 
Songlin Hu$^{1,2}$
\affiliations
$^1$Institute of Information Engineering, Chinese academy of Sciences, Beijing, China\\
$^2$School of Cyber Security, University of Chinese academy of Sciences, Beijing, China\\
$^3$Baidu Inc., Beijing, China
\emails
\{songxiaohui, zangliangjun, suyipeng, hanjizhong, husonglin\}@iie.ac.cn,
wuxing03@baidu.com
}

\begin{document}

\maketitle

\begin{abstract}
    While several state-of-the-art approaches to dialogue state tracking (DST) have shown promising performances on several benchmarks, there is still a significant performance gap between seen slot values (\textit{i.e.}, values that occur in both training set and test set) and unseen ones (values that occur in training set but not in test set). Recently, the copy-mechanism has been widely used in DST models to handle unseen slot values, which copies slot values from user utterance directly. In this paper, we aim to find out the factors that influence the generalization ability of a common copy-mechanism model for DST. Our key observations include: 1) the copy-mechanism tends to memorize values rather than infer them from contexts, which is the primary reason for unsatisfactory generalization performance; 2) greater diversity of slot values in the training set increase the performance on unseen values but slightly decrease the performance on seen values. Moreover, we propose a simple but effective algorithm of data augmentation to train copy-mechanism models, which augments the input dataset by copying user utterances and replacing the real slot values with randomly generated strings. Users could use two hyper-parameters to realize a trade-off between the performances on seen values and unseen ones, as well as a trade-off between overall performance and computational cost. Experimental results on three widely used datasets (WoZ 2.0, DSTC2, and Multi-WoZ 2.0) show the effectiveness of our approach.
    \end{abstract}

    \section{Introduction}

    Task-oriented dialogue system interacts with users in natural language to accomplish tasks such as restaurant reservation or flight booking. The goal of dialogue state tracking is to provide a compact representation of the conversation at each dialogue turn, called \textit{dialogue state}, for the system to decide the next action to take. A typical dialogue state consists of goal of user, action of the current user utterance  (\texttt{inform}, \texttt{request} etc.) and dialogue history \cite{young2013pomdp}. All of these are defined in a particularly designed \textit{ontology} that restricts which slots the system can handle, and the range of values each slot can take. Tracking the user's goal is the focus of this task. To accomplish the tracking task, most DST models take the user's utterance at the current turn, a slot to track and dialogue history as input, output the corresponding value if the user triggers the input slot.
    Considering an example of restaurant reservation, users can \textit{inform} the system some restrictions of their goals (\textit{e.g.}, \texttt{inform(food = thai)}) or \textit{request} further information they want (\textit{e.g.}, \texttt{request(phone number)}) at each turn.
    \begin{table}[h]
        \centering
        \begin{tabular}{|l}
        \textbf{User}:I'm looking for a restaurant that serves thai food. \\
        \texttt{state:inform(food=thai)} \\
        \begin{tabular}[c]{@{}l@{}}\textbf{System}:There are two, one in the west end and one in the\\ centre of town. Do you have preference?\end{tabular} \\
        \begin{tabular}[c]{@{}l@{}}\textbf{User}:The one on the west end, please. Can I have the\\ phone number?\end{tabular} \\
    
        \begin{tabular}[c]{@{}l@{}}\texttt{state:inform(food=thai, area=west)}\end{tabular}\\
            \texttt{request(phone number)}
        \end{tabular}
        \end{table}
    
Having access to an ontology that contains all possible values simplifies the tracking problem in many ways. However, in a real-world dialogue system, it is often impossible to enumerate all possible values for each slot. To reduce the dependence on the ontology, PtrNet\cite{xu2018end} uses the Pointer Network\cite{vinyals2015pointer} to handle the unknown values that are not defined in the ontology, since then the attention-based \textbf{\textit{copy-mechanism}} inspired by Pointer Network become widely used in state-of-the-art DST approaches\cite{ren-etal-2019-scalable,WuTradeDST2019,gao2019dialog}. 
The copy-mechanism based DST models directly copy slot values from the dialogue history, thus reducing the need to pre-define all slot values in the ontology.
    
However, there is still a significant performance gap between seen slot values (\textit{i.e.}, values that occur in both training set and test set) and unseen ones\footnote{In this paper, we use 'unseen' values and 'unknown' values interchangeably.} (\textit{i.e.}, values that occur in only training set but not test set). In this paper, our goal is two-fold: 1) we figure out which factors influence the generalization ability of copy-mechanism based models; 2) we develop a simple but effective algorithm to improve the performance of predicting unseen slot values.

We conduct two experiments to figure out the reason for the poor generalization performance of copy-mechanism based models. In the first experiment, we replace all slot values in either test set or training set with randomly generated values, and we observe that F1-scores drop dramatically in each experimental setting. Consequently, we conclude that the model attempts to memorize the slot values rather than infer them from contexts because their contexts keep the same. In the second experiment, to take full advantage of contexts of slot values, we augment a dataset by copying user utterances several times and replacing all slot values with randomly generated strings. We observe that the generalization performance is positively related to the diversity of slot values (\textit{i.e.}, the number of unique values for slots), and that the performance on unseen values is very close to seen values. Consequently, we conclude that the model learns contextual information for slot values and infers correctly them from their contexts.
    
Based on the above observations, we proposed a simple and effective method of data augmentation to train copy-mechanism models.
The algorithm augments the input dataset by copying user utterances and replacing the real slot values with randomly generated strings. 
Importantly, it could determine the optimal size of a synthetic training set automatically. 
Experimental results show that the synthetic training sets greatly improve the overall performances of copy-mechanism models.
Moreover, users could use two hyper-parameters to realize a trade-off between the performances on seen values and unseen ones, as well as a trade-off between overall performance and computational cost. 
    
The rest of this paper is organized as follows: we first review the recent advances in both DST and the copy mechanism in Section \ref{Related Work}. 
Then, we describe the datasets we used and a general copy-mechanism based model in Section \ref{Datasets and Baseline Model}.
Our data analysis process and key observations are described in Section \ref{Data Analysis and Observations}. We propose our data augmentation approach to DST task and evaluate its performance in Section \ref{Approach}. Finally, we conclude our work and discuss the future work.

\section{Related Works}
\label{Related Work}

\subsection{Dialogue State Tracking}


Deep-learning has recently shown its power to the dialogue state tracking challenges \cite{williams2013dialog,henderson2014second,henderson2014third}. Neural Belief Tracker (NBT) \cite{mrkvsic2016neural} applied representation learning to learn features as opposed to hand-crafting features. PtrNet \cite{xu2018end} aimed to handle unknown values. GLAD \cite{zhong2018global} addressed the problem of extraction of rare slot-value pairs. The number of parameters of these models increased with the number of slots. \cite{ramadan2018large} tried to share parameters across slots, but the model had to iterate all slots and values defined in the ontology at each dialogue turn. \cite{rastogi2017scalable} generated a fixed set of candidate values using a separate SLU module but suffered from error accumulation. TRADE\cite{WuTradeDST2019} was a simple copy-augmented generative model that tracked dialogue states without ontology and enabled zero-shot and few-shot DST in a new domain. \cite{gao2019dialog} used the pretrained language model \textit{BERT}\cite{devlin2018bert} and copy-mechanism to predict explicitly expressed values.

\subsection{Copy-Mechanism}

Copy-mechanism in deep learning is a general concept, which means an output is copied from an input sequence. This idea was first proposed in Pointer Network\cite{vinyals2015pointer}. The Pointer Network is designed to learn the conditional probability of an output sequence with elements that are discrete tokens corresponding to positions in an input sequence. It can address the problems such as sorting variable sized sequences and various combinatorial optimization.
Inspired by Pointer Network, CopyNet\cite{gu2016incorporating} first incorporated copy-mechanism in sequence-to-sequence learning, and the Pointer-Generator Network\cite{see2017get} combined the Pointer Network and a generator to retain the ability to produce novel words.
As mentioned in the introduction, \citeauthor{xu2018end} first reformulated DST problem to take advantage of the flexibility enabled by Pointer Network. 
Then, the copy-mechanism become the most common sub-structure of the DST models, which is used in several state-of-the-art DST models such as TRADE\cite{WuTradeDST2019}, COMER\cite{ren-etal-2019-scalable}, etc. 
DSTRead\cite{gao2019dialog} used a span prediction method proposed in DrQA\cite{chen2017reading} for machine reading task, which is also an application of the copy mechanism.

\section{Datasets, Baseline Model and Evaluation}
\label{Datasets and Baseline Model}

In this section, we first describe the datasets we used, then present a baseline model that implements the copy mechanism widely used in DST models, and finally present evaluating metric.

\subsection{Datasets}

\label{datasets}

Our datasets are extracted from three datasets widely used in DST tasks, \textit{i.e.}, WoZ 2.0\cite{wen2016network}, DSTC2\cite{henderson2014second} and Multi-WoZ 2.0\cite{budzianowski2018multiwoz}.
To test the performance of copy mechanism on unseen slot values, we focus on the slots of which the values are non-enumerable.
Table \ref{tab:slots} lists the slots, the number of slot values, and the number of samples in the following experiments. 

\begin{table}[htpb]
    \centering
\begin{tabular}{c|c|c|c}
\hline
\multirow{2}{*}{datasets} & \multirow{2}{*}{slots} & \multirow{2}{*}{\begin{tabular}[c]{@{}c@{}}values\\ train(total)\end{tabular}} & \multirow{2}{*}{\begin{tabular}[c]{@{}c@{}}samples\\ train(test)\end{tabular}} \\
                          &                        &                                                                                &                                                                                \\ \hline
WoZ                         & food                   & 73(75)                                                                         & 2536(1646)                                                                     \\ \hline
DSTC2                         & food                   & 72(73)                                                                         & 11677(9890)                                                                    \\ \hline
\multirow{10}{*}{Multi}       & hotel-name             & 35(37)                                                                         & \multirow{10}{*}{60027(73720)}                                                 \\ \cline{2-3}
                          & train-destination      & 19(20)                                                                         &                                                                                \\ \cline{2-3}
                          & train-departure        & 23(25)                                                                         &                                                                                \\ \cline{2-3}
                          & attraction-name        & 88(95)                                                                         &                                                                                \\ \cline{2-3}
                          & taxi-destination       & 198(214)                                                                       &                                                                                \\ \cline{2-3}
                          & taxi-departure         & 188(210)                                                                       &                                                                                \\ \cline{2-3}
                          & restaurant-name        & 131(139)                                                                       &                                                                                \\ \cline{2-3}
                          & restaurant-food        & 94(95)                                                                         &                                                                                \\ \cline{2-3}
                          & bus-departure          & 1(1)                                                                           &                                                                                \\ \cline{2-3}
                          & bus-destination        & 1(1)                                                                           &                                                                                \\ \hline
\end{tabular}
    \caption{Slots we use in experiments in WoZ 2.0, DSTC2 and Multi-WoZ datasets. Samples include negative samples for the slot gate.}
    \label{tab:slots}
\end{table}

Each sample corresponds to one dialogue turn, along with a 
\textbf{slot}, its \textbf{active} state, and its corresponding \textbf{value}. An example is as follows:
\begin{table}[h]
        \centering
        \begin{tabular}{|l}
        \textbf{utterance}:I'm looking for a restaurant that serves thai food.\\
        \textbf{slot}: \texttt{food},
        \textbf{active}:\texttt{True},
        \textbf{value}: \texttt{thai}
        \end{tabular}
\end{table}

The active attribute of a slot will be \texttt{True} (corresponding to a positive sample) if the user triggers the slot, otherwise it will be set \texttt{False} (corresponding to a negative sample). 
In the test set, each turn of dialogue will be paired with each slot to reach the convincing results.

As shown in Table \ref{tab:slots}, there are a lot of overlapping values over slots between the training set and test set.
To observe the performance on unseen slot values, we design a function of random string generation to generate new unseen slot values. 
Then, on the basis of the random string generation function, we define another function of generating synthesis datasets.

\subsubsection{Random String Generation}
\label{randstr}
The function $\texttt{randstr(strlen)}$ takes string length as input and output a randomly generated string. Algorithm \ref{alg:randstr} describes its generating process. 
It randomly samples chars from a fix char sequence, we define the char sequence as 26 lowercase letters and 3 spaces\footnote{in this paper we use 3 spaces and \texttt{strlen}=10 to generate more natural multi-words values}.
Spaces are used to generate multi-words values, for example, when \texttt{strlen} is 10, the probability of generating multi-words values is about $1-(26/29)^{10} \approx  0.66$. 

 \begin{algorithm}[tbh]
    \caption{Generate Random String(\texttt{randstr})}
    \label{alg:randstr}
    \textbf{Input}:  length of random string $L$\\
    \textbf{Output}: a random string with length $L$

    \begin{algorithmic}[1] 
        \STATE define a char sequence $S = [a-z, '\quad','\quad','\quad']$.
        \WHILE{True}
        \STATE Generate $L$ random integers from the discrete uniform distribution as indexes $idx$.
        \STATE Concatenate all chars indexed by $idx$ in char sequence $S$ $\rightarrow s$.
        \STATE Remove leading and trailing spaces of $s$.
        \IF {len($s$) == $L$}
            \STATE break
        \ENDIF
        \ENDWHILE
    \STATE \textbf{return} $s$
    \end{algorithmic}
    \end{algorithm}

\subsubsection{Synthesis Dataset Construction}
\label{dc}
 The synthesis dataset construction function $DC(D,n,\theta)$ takes an original dataset as input and outputs a new dataset as follows.
 It first duplicate the data samples in which \texttt{active=True} in $D$ into $n$ copies, get $D'$. For each 
 data sample in $D'$ that contains a active slot, the function replaces the corresponding value with a random value generated by \texttt{randstr}, with a probability of $\theta$.
 Specially, all values in set will be replaced if $\theta=1$, and $DC(D,1,0)$ keep
 the input dataset unchanged.

\subsection{Baseline Model}
\label{Baseline Model}

\begin{figure}[tp]
    \centering
    \includegraphics[width=0.48\textwidth]{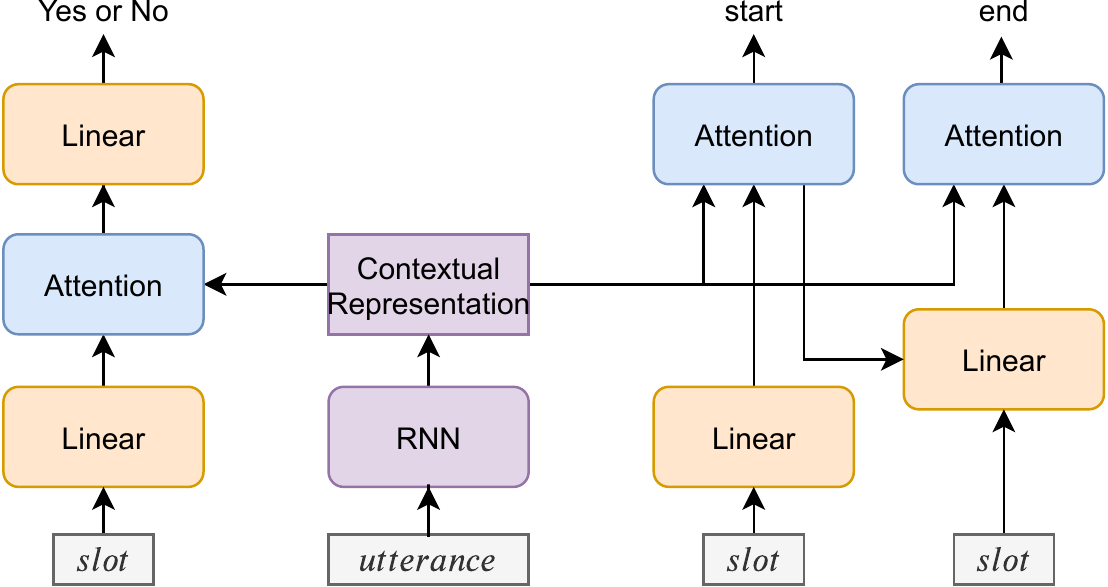}
    \caption{The architecture of our baseline model. The inputs are embedding of slot, word embeddings of user and system utterance($W_1, W_2, \cdots, W_n$). It is a common sub-structure in SOTA  DST models.}
    \label{baseline}
\end{figure}

Without loss of generality, we implement a basic copy-mechanism based model. 
The model takes an utterance $U$ and a slot $s$ as input, and output whether $s$ is active and the positions of its corresponding value if $s$ is active. 
The model consists of three important components: \texttt{utterance encoder}, \texttt{attention-based copy-mechanism}, and \texttt{slot gate}.
Its architecture is presented in Figure \ref{baseline}.

\subsubsection{Utterance Encoder}

For utterance $U$, we simply concatenate user and system utterance ($U^{usr}$ and $U^{sys}$) by a particular symbol ${<}$\texttt{usr}${>}$.
\begin{equation}
  U=U^{sys} \oplus {<}\texttt{usr}{>} \oplus U^{usr}
\end{equation}
\noindent We use a bidirectional LSTM \cite{hochreiter1997long} to get the contextual representation $H^{P}$ of $U$.
\begin{equation}
  H^{P} = \mathrm{BiLSIM}^{P}(U_{emb})\in \mathbb{R}^{n \times d_h}
\end{equation}
\noindent where $n$ denotes the number of words in $U$, $U_{emb}\in \mathbb{R}^{n \times d_{emb}}$ the word embeddings of $U$, $d_{emb}$ the dimension of embeddings, and $d_h$ the dimension of LSTM hidden states.

\subsubsection{Attention-based Copy-Mechanism}

We define an attention function $Attn(Q,V)$ to calculate attention scores between query feature $Q\in \mathbb{R}^{d_{emb}}$  and context feature $V\in \mathbb{R}^{n\times d_h}$. The computing process is as follows: 1) do a linear transform for both $Q$ and $V$, 2) use the dot product result as attention scores, 3) normalize through \texttt{softmax} function.
\begin{align}
    Q' = QW_{q},
    V' & = VW_{v},
    \alpha_{i} = Q'V'_{i}\\
    scores_i &= \exp \alpha_i/\sum_j^{n} \exp\alpha_j\\
    contexts &= \sum_i^n scores_i V'_i
\end{align}

We use a single linear layer($\mathrm{Linear}(X) = WX+b$) to encode slots, and then use the attention function defined above to calculate both start and end positions of its value.

\begin{align}
    s_{enc} &= \mathrm{Linear}_{slot}(s_{emb})\\
    p_{enc} &= \mathrm{Linear}_{p}(s_{enc})\\
    contexts^{p}, scores^p &= Attn_{span}(p_{enc}, H^{P})\\
    q_{enc} &= \mathrm{Linear}_{q}(s_{enc} \oplus contexts^{p})\\
    contexts^{q}, scores^q &= Attn_{span}(q_{enc}, H^{P})\\
    start &= \mathop{\arg\max}\limits_{j} scores^{p}_{j}\\
    end &= \mathop{\arg\max}\limits_{j} scores^{q}_{j}
\end{align}

\subsubsection{Slot Gate}

A binary classifier is used to determine whether a slot is triggered by a user or not, where the single linear layer $\mathrm{Linear}_{cls2}$ produces a probability over $[0,1]$ based on attention contexts.
A slot is triggered if $cls_{result} > 0.5$.

\begin{align}
    s_{cls} &= \mathrm{Linear}_{cls1}(s_{emb})\\
    contexts^{cls}, scores^{cls} &= Attn_{cls}(s_{cls}, H^{P})\\
    \alpha_{cls} &= \mathrm{Linear}_{cls2}(contexts^{cls})\\
    cls_{result} &= \mathrm{sigmoid}(\alpha_{cls})
\end{align}

\subsubsection{Implementation Details}
\label{details}

To enhance the effectiveness of the experiments, all experiments in this paper share the same settings. 
We use randomly initialized word embeddings of dimension 300 with dropout\cite{JMLR:v15:srivastava14a} rate 0.5. The model is trained with Adam\cite{kingma2014adam} optimizer and the learning rate is 1e-4. 

In all experiments, we train the baseline model for 80 epochs on WoZ dataset, 32 epochs on DSTC2 dataset and 32 epochs on Multi-WoZ dataset to guarantee good convergence. At each epoch, we evaluate the model on the dev set and save the checkpoint and select the best to get the final result on the test set when the training process completes. All experiments are conducted on a single NVIDIA RTX 2080Ti GPU.

\subsection{Evaluation Metric}
\label{Evaluation Metric}
We use F1 scores as the primary metric in all experiments in this paper. Specifically, if a slot is determined to be active by the slot gate and the model predicts the correct value (both the start and end positions) for the slot, then it is a true positive sample. If the slot gate outputs \texttt{False}, then the extracted value will be ignored. 

\section{Data Analysis and Observations}
\label{Data Analysis and Observations}
In this Section, we analyze the factors that influence the generalization performance of the baseline model.

\subsection{The model tends to memorize values}
\label{Model tend to memorize values}
    As shown in TRADE\cite{WuTradeDST2019}, the slots that share similar values or have correlated values have similar embeddings. We guess that the attention based copy-mechanism is inclined to memorize values that appear in the train set rather than infer them from contexts. 
    To verify our argument, we conduct four groups of experiments using original and synthetic datasets. 
    The experimental results are presented in Table \ref{tab:memorizevalue}, where the synthetic training/test sets are built using the dataset construction function $DC(D,1,1)$ (see definition in section \ref{dc}) under constraint of sharing the same set of randomly generated values.

    \begin{table}[h]
        \centering
        \begin{tabular}{c|ccc}
            \hline
             & dataset & \begin{tabular}[c]{@{}c@{}}original\\ test\end{tabular} & \begin{tabular}[c]{@{}c@{}}synthetic\\ test\end{tabular} \\ \hline
            \multirow{3}{*}{\begin{tabular}[c]{@{}c@{}}original\\ train\end{tabular}} & WoZ2.0 & 0.9241 & 0.5801 \\
             & DSTC2 & 0.9850 & 0.3231 \\
             & MultiWoZ & 0.9079 & 0.4412 \\ \hline
            \multirow{3}{*}{\begin{tabular}[c]{@{}c@{}}synthetic\\ train\end{tabular}} & WoZ2.0 & 0.5283 & 0.9586 \\
             & DSTC2 & 0.2385 & 0.9863 \\
             & Multi-WoZ & 0.3631 & 0.8837 \\ \hline
            \end{tabular}
        \caption{F1 scores on three datasets. The synthetic training/test sets are constructed using $DC(D,1,1)$ with the same set of slot values.}
        \label{tab:memorizevalue}
    \end{table}
    
 As shown in Table \ref{tab:memorizevalue}, the model perform well when both training and test set are original or synthetic, but perform poorly when one is original and the other is synthetic.
 The former corresponds to the case of seen slot values, and the latter  corresponds to the case of unseen slot values.
 Since we do not change the contexts of slot values, we conclude that the model attempts to memorize the slot values rather than infer them from contexts.
 Interestingly, the model shows pretty good performances in the case that training/test set are both synthetic, which confirms that it is reasonable to use random strings to augment datasets.

\subsection{Greater diversity of values brings better generalization}

We have argued that the model tends to memorize the values that appear in the train set. In other words, the model pay little attention to context information of slot values, which is a very useful feature for the model to recognize the correct value for a slot.
Hence, we would like to make full use of context information of slot values. 
Intuitively, high diversity of slot values make the model more difficult to learn from slot values, and duplicated contexts make the model easier to learn from less varied contexts.
Consequently, an interesting question is: 
\begin{table}[h]
        \centering
        \begin{tabular}{|p{0.4\textwidth}}
        \textit{If we greatly increase the diversity of slot values and duplicate contexts in the train set, will the model prefer inferring slot values from slot contexts?}
        \end{tabular}
\end{table}

\subsubsection{Greater diversity of values improves  generalization}

\begin{figure}[htp]
    \centering
    \includegraphics[width=0.4\textwidth]{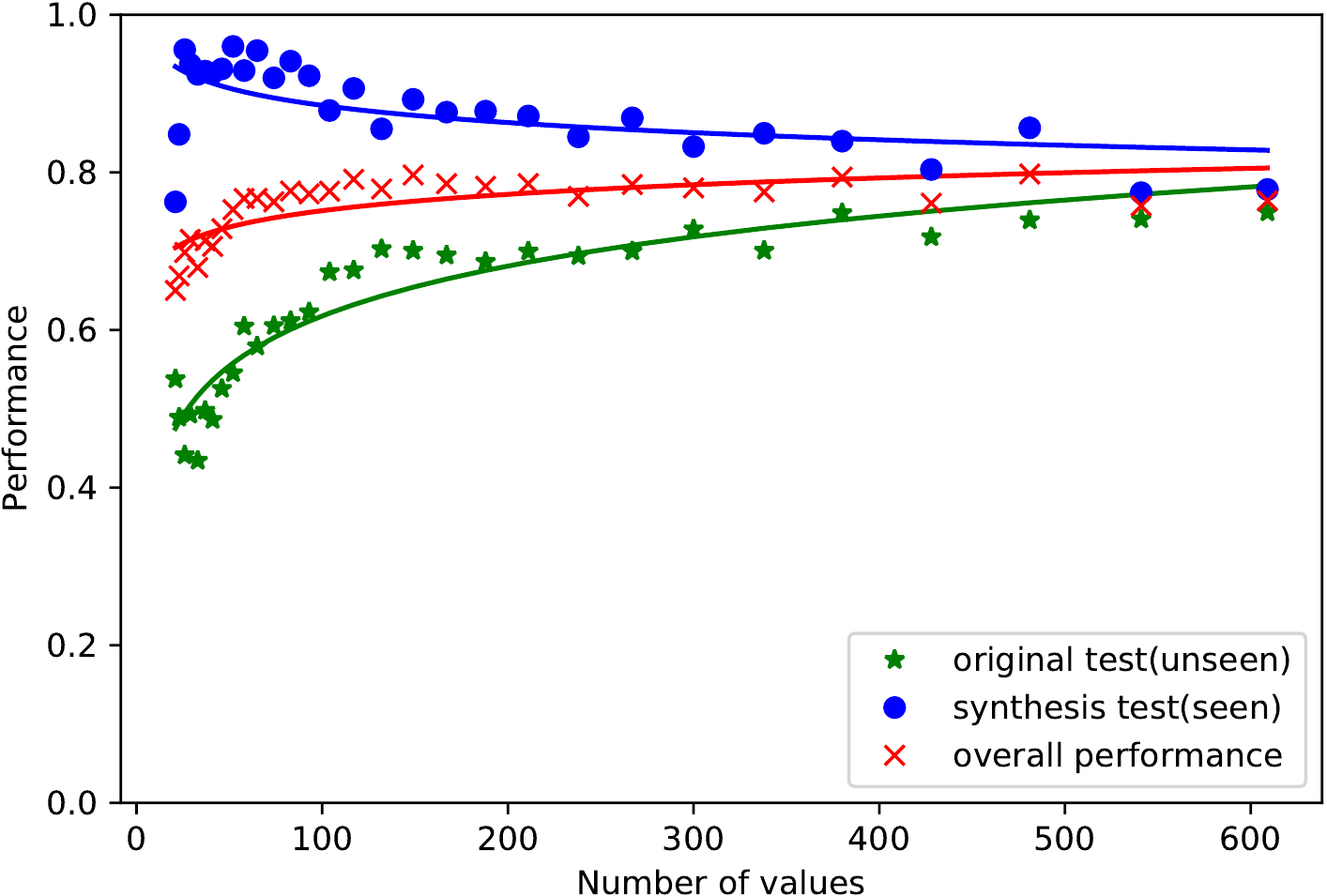}
    \caption{Performance on dataset constructed using
    $DC(\mathrm{WoZ},1,1)$ with different size of values set, values in training and modified test set are all randomly generated from the same values set,\textit{ all values in original test set are unseen values}. Overall performance in the figure is the mean of F1 score at the two test sets. Experiments also conducted on DSTC2 and Multi-WoZ dataset and got similar results.}
    \label{value_num}
\end{figure}

We could adjust the diversity of slot values by controlling the number of slot values in the training set.
Let $N$ be the number of all slot values (may duplicate) appeared in the training set. 
We set the numbers of unique values with $[20, 20\times \alpha^{1},\cdots,20\times \alpha^{29}=N]$ (\textit{i.e.}, totally 30 numbers), where $20$ and $N$ is the minimum and maximum number of unique slot values respectively. 
We sample in this way to present a clear trend for small number of values. 
For each number, we first use the function of random string generation (see algorithm \ref{alg:randstr}) to produce a new set of slot values, and then use the function $DC(D,n= 1,\theta=1)$ of dataset construction to produce a synthetic training set and a synthetic test set with the same set of slot values. 
In this case, all values in the synthetic test set could be found in the synthetic training set, and all values in the original test set are completely unseen for the synthetic training set.

As shown in Figure \ref{value_num}, the F1-scores of original test set (\textit{i.e.}, unseen values) and average F1-scores increase rapidly with increasing number of unique slot values. Hence, we conclude that the diversity of values are positively correlated to model's generalization ability.

\begin{figure}[bhp]
    \centering
    \includegraphics[width=0.4\textwidth]{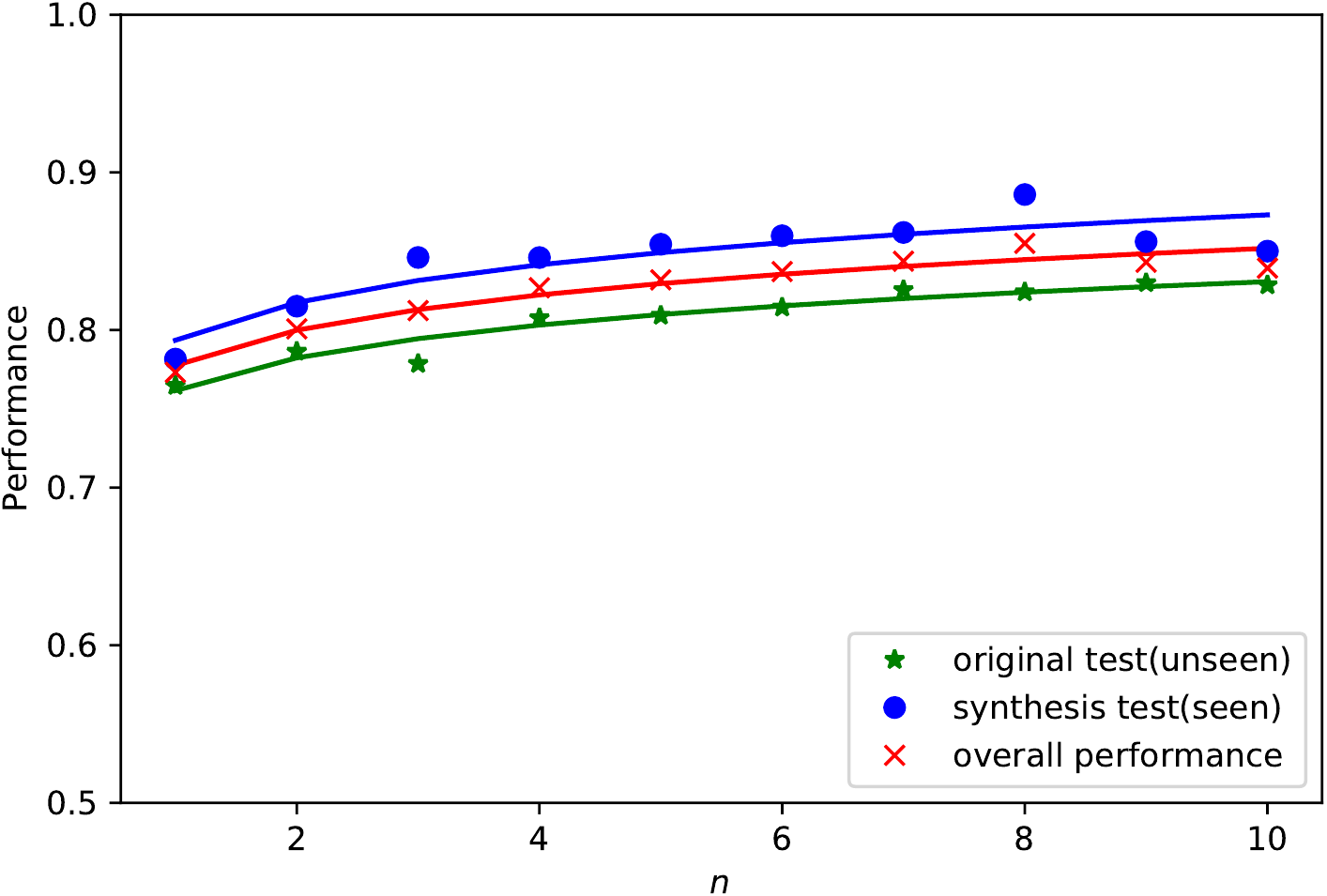}
    \caption{Performance on dataset constructed by function $DC(\mathrm{WoZ},n,1)$, $n \in [1..10]$. \textbf{Note} that in this figure, we didn't control the size
    of values set, the size of values set increased as $n$ increased.}
    \label{aug_time}
\end{figure}

\subsubsection{More samples further improves generalization}

In Figure \ref{value_num}, the number of different values is less than or equal to the number of samples that contains active slots. What will happen if we continue to increase the number of different values? In this experiment, we use the function $DC(D,n,\theta=1)$ to produce synthetic training/test set, where $n$ is taken from $[1..10]$ and  all generated values are different.

As shown in Figure \ref{aug_time}, the F1-scores of original/synthetic test sets and average F1-scores increase with increasing number of unique slot values. Hence, we conclude that the number $n$ of copies are positively correlated to model's generalization ability.


To sum up, we have similar conclusion from Figure \ref{value_num} and Figure \ref{aug_time}. That is, with increasing number of slot values, the performance on unseen values and the overall performance increase substantially but the increasing speed gets slower. 
Therefore, increasing the diversity of values of the training set to some extent is a feasible way to improve the model’s generalization.
\label{seenandunseen}
In addition, we can find that the performance on seen values (\textit{i.e.}, synthetic test set) is always better than that on unseen values (\textit{i.e.}, original test set). 

\section{Data Augmentation for DST}
\label{Approach}
In this section, we first propose a simple data augmentation algorithm based on the conclusions in Section \ref{Data Analysis and Observations}, and then evaluate our algorithm experimentally.


  \begin{algorithm}[htbp]
    \caption{Data Augmentation Algorithm}
    \label{alg:algorithm}
    \textbf{Input}:  datasets $D$, $T_s$, $T_u$, randomly initialized tracker $T$\\
    \textbf{Hyperparameters}:  the proportion of the new value $\theta$, eps=$\epsilon$\\
    \textbf{Output}: the times $n$ that copies original dataset, the tracker $T$ trained on $DC(D,n,\theta)$, the performance $P$ of $T$.

    \begin{algorithmic}[1] 
        \STATE Let the number of copies $t=1$, performance $res$=[]
        \STATE train $T$ on $D$, and get an overall performance $P$ on $T_s$ and $T_u$ follow steps in section \ref{details}\label{step3}
        \STATE $res \leftarrow P$
    \WHILE {True}
    \STATE re-initialize the tracker $T$
    \STATE $D'=DC(D,n,\theta)$
    \STATE train $T$ on $D'$, get the overall performance $P$ on $T_s$ and $T_u$ the same as step \ref{step3}
    \STATE $res \leftarrow P$
    \STATE $n = n * 2$
    \IF {len($res$) $<$ 3}
        \STATE continue
    \ENDIF
    \IF {$res$[-1] - $res$[-2] $< \epsilon$ and $res$[-2] - $res$[-3] $< \epsilon$}
        \STATE break
    \ENDIF
    \ENDWHILE
    \STATE \textbf{return} $n, T, \max(res)$
    \end{algorithmic}
\end{algorithm}



\subsection{Approach}

Given a training set $D$, a test set $T_s$ containing all seen values and another test set $T_u$ containing all unseen values, our algorithm outputs the optimal number $n$ for the optimization problem as follows:
    \begin{equation}
        \arg\min_{n\geq 2} \{P_\theta(2n)-P_\theta(n)\leq \epsilon \texttt{ and } P_\theta(n)-P_\theta(n/2)\leq \epsilon\}
    \end{equation}
where  $P_\theta$ is the average of the performances on $T_s$ and $T_u$.

There are two hyper-parameters $\theta$ and $\epsilon$ in our algorithm. 
Firstly, $\theta$ is used to control the proportion of unseen values when we generate a training set using the function $DC(D,n,\theta)$.
Users can use $\theta$ to trade-off the performances between seen values and unseen ones.
For example, a user may set $\theta$ with 0.5 when he believe that seen values and unseen values are equally important.
Secondly, $\epsilon$ is used to control the terminating condition of the algorithm.
Even though the overall performance is positively related to the number $n$ of copies in $DC(D,n,\theta)$, the computational cost also increases when the size of training set increases. 
Users can use $\epsilon$ to trade-off between performance and computational cost. 

\textbf{This problem can be defined to find a point in Figure \ref{aug_time} that has the best overall performance.} 
We use the \textbf{doubling} method to search the best $\mathbf{n}$ to construct the training set within an acceptable computational cost. 
Algorithm \ref{alg:algorithm} presents the details of this process.

\subsection{Performance}
\label{Evaluation}
We evaluate our data augmentation algorithm on three datasets: WoZ 2.0, DSTC2 and Multi-WoZ. 
We focus on slots that have non-enumerable values (see Table \ref{tab:slots}) and present experimental results in Table \ref{tab:results}. 
Here we use $\theta=0.5$ to balance the ratio between seen and unseen values, and set the precision $\epsilon$ to 0.01 to get best overall performance
  within limited computational resources.

From Table \ref{tab:results} we can find that, with our data augmentation algorithm, the performance of the baseline model on unseen values improves significantly, and the performance on seen values decrease slightly.

 \begin{table}[h]
    \centering
    \begin{tabular}{|l|l|l|l|}
    \hline
    Dataset & Model & seen & unseen \\ \hline
    \multirow{2}{*}{WoZ 2.0} & baseline & 0.9241 & 0.5801 \\ \cline{2-4} 
     & +DA(n=64) & 0.9073\tiny ($\downarrow$1.7\%) & 0.8818\tiny (\textbf{$\uparrow$30.17\%}) \\ \hline
    \multirow{2}{*}{DSTC2} & baseline & 0.9850 & 0.3231 \\ \cline{2-4} 
     & +DA(n=32) & 0.9711\tiny($\downarrow$1.39\%) & 0.9591\tiny(\textbf{$\uparrow$63.6\%}) \\ \hline
    \multirow{2}{*}{Multi-WoZ} & baseline & 0.9079 & 0.4412 \\ \cline{2-4} 
     & +DA(n=32) & 0.9032\tiny($\downarrow$0.65\%) & 0.8847\tiny(\textbf{$\uparrow$44.35\%}) \\ \hline
    \end{tabular}
    \caption{F1 scores of our data augmentation(DA) approach on three datasets' test set(seen) and modified test set(unseen), with $\theta=0.5$ and $\epsilon=0.01$. And $n$ is the number of copies in the dataset construct function $DC(D,n,\theta)$  that gains the best overall performance in Algorithm \ref{alg:algorithm}.}
    \label{tab:results}
    \end{table}

\subsection{Hyperparametric Analysis}

\begin{figure}[h]
    \centering
    \includegraphics[width=0.4\textwidth]{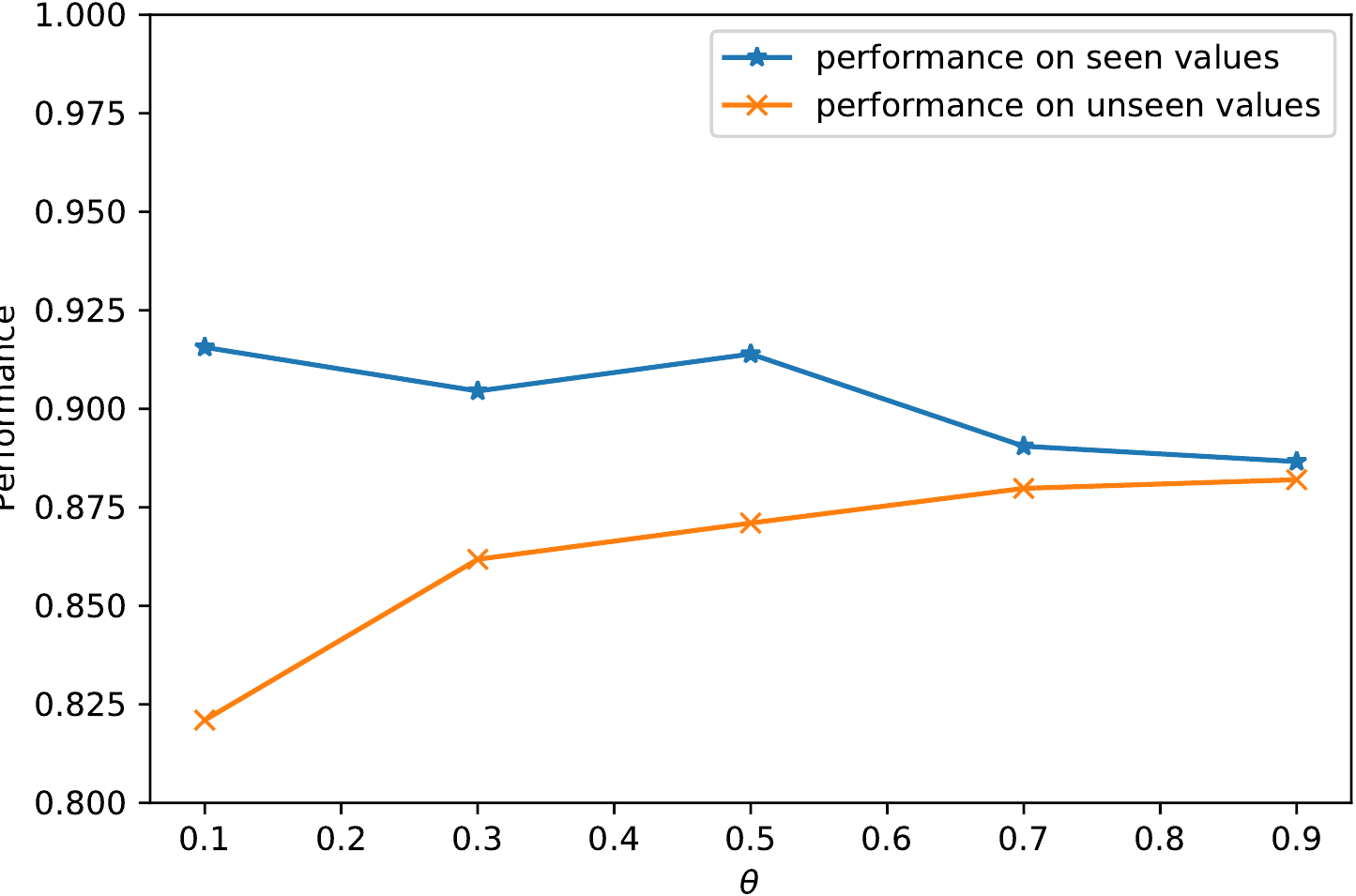}
    \caption{The effect of $\theta$ on WoZ dataset with $\epsilon=0.01$.}
    \label{theta}
\end{figure}

We introduce two hyperparameters in our data augmentation algorithm, $\theta$ and 
$\epsilon$. $\theta$ is used in the dataset construct function $DC(D,n,\theta)$ to control the proportion of new generated values in the train set, and $\epsilon$ is used in the terminate condition of the doubling searching to trade off the performance and computational cost. The relation between performance and $\theta$ is 
shown in Figure \ref{theta}, and the effect of $\epsilon$ is shown in Table \ref{tab:eps}.

In Figure \ref{theta}, we apply different $\theta$ to our algorithm, and record the performance $P_s$ and $P_u$ for each $\theta$. 
We can find that \textbf{a larger $\theta$ increases the performance on unseen values but decreases the performance on seen values}. 
As we described in section \ref{seenandunseen}, performance on seen values always better than that on unseen values. 
This figure confirm that the proportion of new generated values could be used to balance the performance on seen and unseen values.

As the number of search steps increases, the performance gain gets smaller, which can be concluded from Figure \ref{aug_time}. 
From table \ref{tab:eps}, we could observe that \textbf{a small $\epsilon$ leads to more searching steps and better performance, while a large $\epsilon$ means lower computational cost and slightly worse performance}.
Thus, we could use $\epsilon$ to realize a trade-off between performance and computational cost.



\begin{table}[h]
    \centering
    \begin{tabular}{c|c|c}
    \hline
    $\epsilon$ & search steps & overall performance \\ \hline
    0.01 & 7(t=64) & 0.8946 \\ \hline
    0.02 & 6(t=32) & 0.8871 \\ \hline
    0.03 & 6(t=32) & 0.8871 \\ \hline
    0.04 & 4(t=8) & 0.8443 \\ \hline
    0.05 & 3(t=4) & 0.8443 \\ \hline
    \end{tabular}
    \caption{The effect of the $\epsilon$ on the WoZ dataset with $\theta=0.5$.}
    \label{tab:eps}
    \end{table}

\section{Conclusion and Future Work}
\label{Conclusion and Future Work}

This paper focuses on the problem of improving generalization ability of copy-mechanism based models for DST task, especially for the slots that have non-enumerable values. 
Our conclusions include:
Firstly, the copy-mechanism model attempts to memorize values rather than infer them from contexts, which is the crucial reason for unsatisfactory generalization performance. 
Secondly, the model's generalization improves dramatically with increasing diversity of slot values. 
Thirdly, data augmentation for copy-mechanism models is feasible and effective in improving generalization of these models.

In future work, we can use the character-level features of the
slot values since the values for the same slot often share somehow similar spelling. 
For example, an address or location often contains uppercase letters, and a slot about time often consists of Arabic numerals and symbol ':'. 
In addition, the effect of contexts' 
diversity remains unexplored, which may help reduce computational cost further.
\bibliographystyle{named}
\bibliography{ijcai20}

\end{document}